# Neurosymbolic hybrid approach to driver collision warning

Kyongsik Yun*[a], Thomas Lu[a], Alexander Huyen[a], Patrick Hammer[b], Pei Wang[b]

[a] Jet Propulsion Laboratory, California Institute of Technology, 4800 Oak Grove Dr, Pasadena, CA 91109, [b] Temple University, 1801 N Broad St, Philadelphia, PA 19122

## ABSTRACT

There are two main algorithmic approaches to autonomous driving systems: (1) An end-to-end system in which a single deep neural network learns to map sensory input directly into appropriate warning and driving responses. (2) A mediated hybrid recognition system in which a system is created by combining independent modules that detect each semantic feature. While some researchers believe that deep learning can solve any problem, others believe that a more engineered and symbolic approach is needed to cope with complex environments with less data. Deep learning alone has achieved state-of-the-art results in many areas, from complex gameplay to predicting protein structures. In particular, in image classification and recognition, deep learning models have achieved accuracies as high as humans. But sometimes it can be very difficult to debug if the deep learning model doesn't work. Deep learning models can be vulnerable and are very sensitive to changes in data distribution. Generalization can be problematic. It's usually hard to prove why it works or doesn't. Deep learning models can also be vulnerable to adversarial attacks. Here, we combine deep learning-based object recognition and tracking with an adaptive neurosymbolic network agent, called the Non-Axiomatic Reasoning System (NARS), that can adapt to its environment by building concepts based on perceptual sequences. We achieved an improved intersection-over-union (IOU) object recognition performance of 0.65 in the adaptive retraining model compared to IOU 0.31 in the COCO data pre-trained model. We improved the object detection limits using RADAR sensors in a simulated environment, and demonstrated the weaving car detection capability by combining deep learning-based object detection and tracking with a neurosymbolic model.

## 1. INTRODUCTION

In the past few years, deep learning algorithms have achieved state-of-the-art results in many areas including computer vision and natural language understanding [1–5]. In particular, in image classification and object recognition, deep learning models have achieved human-level accuracy [6–8]. Despite the unprecedented performance of deep learning, there are still some shortcomings. The most important aspect is that deep learning systems lack explainability and reliability when they do not perform well. The complexity of deep learning systems makes it difficult to pinpoint the cause of the problem. Therefore, in the autonomous driving industry, major companies, including Tesla (HydraNet), Waymo (ChauffeurNet) and Ford/Volkswagen Group (Argo AI), are using modular systems that combine multiple independent modules rather than end-to-end for autonomous driving functions. The "divide and conquer" approach helped us identify and fix problems while solving complex problems.

Another challenge in deep learning is its high sensitivity to a given data set, which prevents generalization. In the ideal situation where the given training data distribution covers the entire edge case, this should be fine. However, it is very difficult to collect sufficient data from the edge cases, especially in the autonomous driving industry, such as crashes, unexpected road conditions, etc.. This is where neurosymbolic hybrid systems can help. Deep learning systems combined with well-defined logic-based systems that establish physical and logical boundary conditions will accelerate the performance of autonomous driving systems even in edge cases. Moreover, a hybrid system would be less susceptible to biased data distribution or adversarial attacks as the system would be controlled and constrained by the values of the logic and the rules of the road, rather than relying solely on the given training data [9,10].

To address the limitations of deep learning, we combined deep learning-based object recognition and tracking with an adaptive neurosymbolic network agent called Non-Axiomatic Reasoning System (NARS) [11]. The specific topic we were trying to address in this study was driver crash warning, especially for first responders, including emergency vehicles and police vehicles. Emergency car accidents in the United States cost $35 billion annually [12]. Collision fatalities are 4.8 times higher than the national average for first responders. Police officers have twice the rate of car accidents per million vehicles they drive than the general public.

Here, we aim to develop and demonstrate an AI assistant that enables first responders to drive safely to avoid accidents in and around high traffic. We implement a hybrid deep learning and neurosymbolic algorithm to provide explainable and trustable autonomy techniques for driver safety recommendations in time-critical first responder operations.

## 2. CARLA DRIVING SIMULATION ENVIRONMENT

CARLA is an open source driving simulation software that can simulate various driving environments and crash scenarios [13]. CARLA supports the development, training and validation of autonomous driving systems. The CARLA system receives driving scenarios, including user-controlled inputs and predefined multi-vehicle movement scripts, and outputs simulated videos and images, as well as various telemetry data. Also in this study, several virtual map creation tools were used, including OpenStreetMap [14] for editable world maps, OpenDrive [15] for an open format specification that describes the logic of road networks, and RoadRunner [16] from MathWorks for designing 3D scenes for automated driving simulations.

We designed a CARLA server and client environment, and a neurosymbolic hybrid system (Fig. 1). Each client sends user actions and scripted scenario data to the server. And the CARLA server synchronizes with multiple clients to send visible and depth camera streams, dynamic vision sensor camera streams, Global Navigation Satellite System (GNSS) telemetry, accelerometer, gyroscope and compass, LIDAR and radar telemetry data to clients. The CARLA server also provides basic signal processing, including collision detection, obstacle detection, semantic LIDAR, and semantic segmentation of objects in the scene.

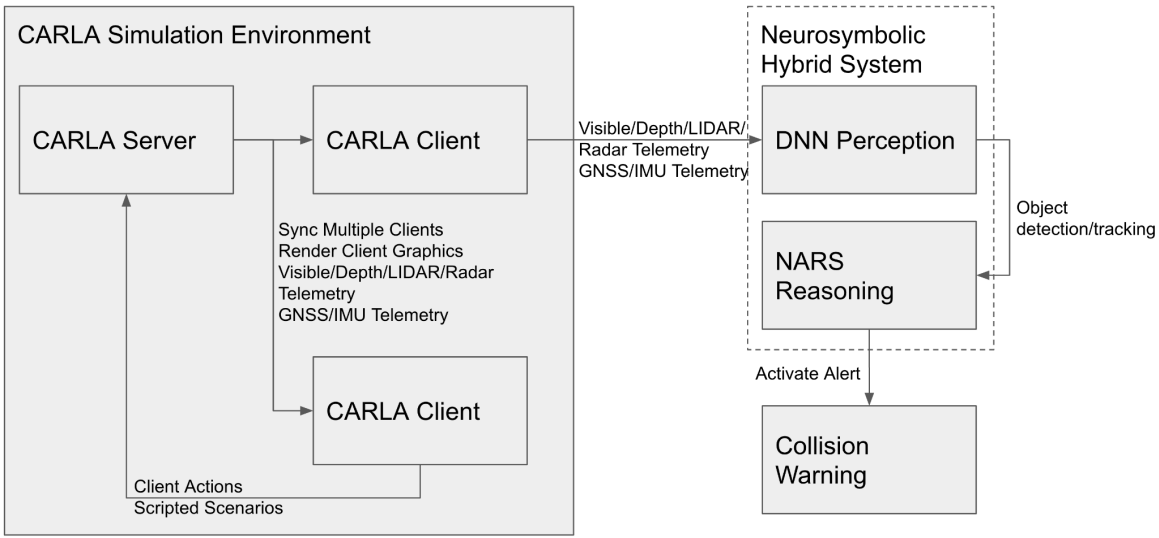

Figure 1. CARLA simulation environment, Neurosymbolic Hybrid System, including deep neural network-based perception for object detection and tracking, and Non-Axiomatic Reasoning System (NARS) for collision warning.

The CARLA simulator uses a Python development environment to provide an easy interface with external modules and algorithms. Deep learning-based object detection and tracking algorithms have real-time access to CARLA's video, LIDAR, and radar data streams, allowing object detection and tracking systems to seamlessly process the data to generate the information needed for the adaptive neurosymbolic system (NARS). Finally, our system provides information about driver priority and collision warning via NARS (Figure 1).

We created various scenarios in the CARLA environment. First responder vehicles can be manually controlled as ego vehicles. Non-player character (NPC) vehicles drive, brake and follow waypoints according to a scenario. We set pass or fail criteria for each scenario to improve the algorithm and avoid collisions. The first scenario was intersection navigation assistance. A first responder vehicle approaches the intersection without slowing down. Another civilian vehicle is speeding through the intersection and there is a high risk of colliding with the first responder vehicle. When the first responder vehicle approaches an intersection on the street: NARS searches the crash history database to find information about past crashes at this intersection as a reference for risk analysis. NARS also checks traffic patterns to see if parameters in the crash history match current traffic patterns. NARS calculates the risk, warns first responders ahead of potential collisions through intersections, and warns drivers to slow down to a safe speed.

The second scenario is stationary vehicle assistance. A first responder vehicle is parked behind civilian vehicles on the shoulder. If a first responder vehicle attempts to park on the shoulder: NARS searches the crash history database to find the crash history of that road as a reference for risk analysis. After matching the time and location to the crash record, NARS calculates the risk, warns of a potential crash at that location, and advises the driver to move to a safer location. NARS can provide 360-degree situational awareness for shoulder-parked first responder vehicles. If a nearby civilian vehicle exhibits dangerous behavior, the system will warn the first responder of the risk of a collision in advance.

CARLA ran on Ubuntu OS 18.04 with AWS (NVIDIA T4 GPU instance). It took roughly 3 hours to plan and create a new scenario, and 0.5 to an hour to modify an existing one.

## 3. DEEP LEARNING-BASED OBJECT DETECTION AND TRACKING

For deep learning-based object detection and tracking, the first step is to generate a dataset for training and testing the model. We generated training data in the CARLA environment. The meaning of generating training data is to label objects with bounding boxes. The first labeling step is to create a bounding box for the vehicle. You can get very accurate bounding boxes by pulling information from the CARLA simulation world itself. All vehicles in the CARLA environment were extracted from each video frame. Then we removed the vehicle that was too far away (200 meters or more). We've also gone through the process of removing vehicles that are obstructing other objects.

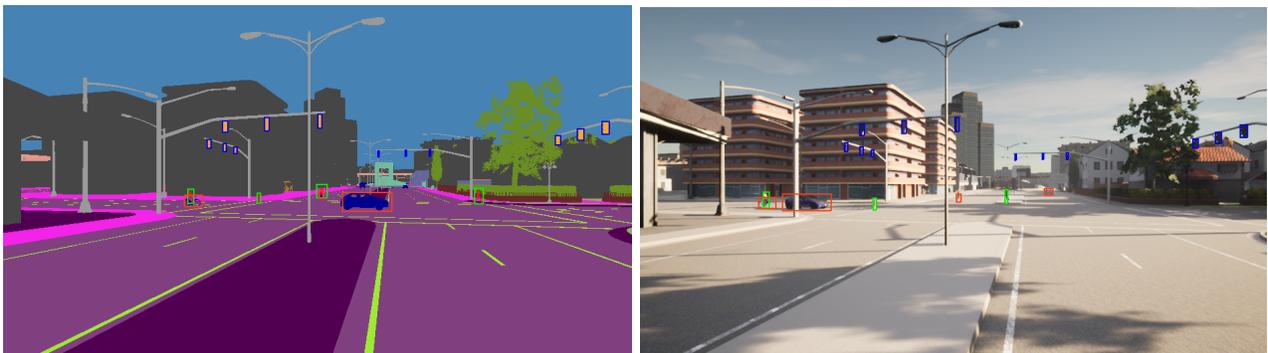

Figure 2. Semantic segmentation of the CARLA simulation environment. We used the segmentation data to localize each object and plotted bounding boxes for subsequent deep learning model training. This was done automatically to generate large amounts of training data in various driving environments..

Because CARLA does not provide the same kind of information for non-vehicle actors as vehicles, new data extraction methods are required to obtain bounding boxes for all objects other than vehicles. We overlayed the semantic segmentation image provided by CARLA, and used a machine learning clustering algorithm (DBSCAN [17,18]) to find each cluster of similar color. This allowed us to extract bounding boxes for non-vehicle objects.

The objects we target are pedestrians, traffic lights, traffic signs and vehicles. We generated data containing a total of 4,480 training images and 1,120 test images. Image resolution is important for recognizing distant objects, so we saved the image at the highest possible resolution of 1920 x 1080px. We also organized the training data to be as generalizable as possible without overfitting the model training by varying weather conditions and times.

We collected training data in a CARLA environment for object detection and compared the performance of each deep neural network to determine the best model for object detection in our environment. Then, we formatted the model output for integration with the NARS adaptive neurosymbolic network for alerting first responders. We also compared the accuracy of various CARLA sensors, including vision, radar, LIDAR, and depth cameras for object tracking. The CARLA sensor was then integrated with object detection and NARS systems. Finally, the safety of the given scenario was evaluated.

## 4. ADAPTIVE NEUROSYMBOLIC NETWORK

Non-Axiomatic Reasoning System (NARS) is an intelligent model that can perform learning and reasoning in various domains [11,19]. NARS is a system that can draw tentative conclusions from available knowledge and make inferences only with uncertain and incomplete information. NARS can respond quickly even when there is not enough time to thoroughly evaluate all possibilities. NARS can not only directly input human logic, but also learn from its own success and failure experiences by interacting with the test environment.

In this study, NARS takes as input the output of the driving simulator (CARLA) and the output of object detectors and trackers. NARS predicts risk by combining input from multiple forms with background knowledge. A predicted risk triggers a warning message to the user. NARS also provides recommendations for avoiding or reducing predicted risks. All conclusions generated by the system are attached with confidence and priority values, and the decision logic can be transparently described according to the derivation process.

Each piece of generative knowledge has a truth value and a priority value, which can be triggered by the example conditions below. 1. A warning is issued when a first responder vehicle enters an intersection at a red light and another vehicle is rapidly approaching from the left or right. 2. If a first responder vehicle is moving, and there is a pedestrian in front, an alert is issued. 3. If a first responder vehicle is crossing an intersection and another first responder vehicle is expected to cross from the left or right, an alert is issued.

Knowledge can be acquired in several ways. They can be provided directly by human experts, or they can be extracted from traffic regulations and laws. We can also acquire knowledge derived from input data or derived from other knowledge. The performance of the NARS system improves over time and can adapt to changes in the environment. NARS receives vehicle type, location, speed and direction information. The knowledge database also collects background knowledge about the missions and notifications to be generated. The output of NARS is the relevant notifications and alerts for first responders. A knowledge base is mission-related knowledge of an intended use case. The knowledge base also includes environmental information, including weather conditions, locations of other first responders, and target tasks.

Moreover, NARS can search the local automobile accident history database and utilize it for risk analysis. It warns of potential collisions in the area and advises drivers. At each stage, the operation of the NARS's experience buffer (judgment, questioning, or goal setting) interacts with the concept of memory according to its rules. Tasks and concepts are selected probabilistically based on their priorities. Factors affecting the priority of an item include its quality, past usefulness, and relevance to its current context. NARS processes many concepts in parallel.

A collection of concepts stored in a NARS-specific data structure is called a "bag" and is actively maintained and prioritized by the system. Concepts with higher priority are more likely to be selected. Concepts store beliefs (i.e., knowledge of the system) that are used for reasoning. All concepts in NARS are fluid. Its meaning is determined not by references or definitions, but by experienced relationships with other concepts. The meaning of the concept is changed as a function of the new experience acquired.

**BACKGROUND KNOWLEDGE:**
//If there is a car approaching from either side, activate alert to prevent a crash
<(<#1 --> ([approaching] & car)> &/ <(*,{SELF}) --> ^alert>) =/> (--,<{SELF} --> [crash]>)>.

**EXAMPLE:**
//Detected object **obj12 is** a **car**
<{obj12} --> car>. :|:
//**obj12 is approaching**
<{obj12} --> [approaching]>. :|:
//(by *compositional inference*)
//**obj12 is** an **approaching ca**
<{obj12} --> ([approaching] & car)>. :|:
//Process Goal: Do not let TruePAL crash
(--,<{SELF} --> [crash]>)!
//activate alert according to background knowledge
**NARS: activate ^alert**

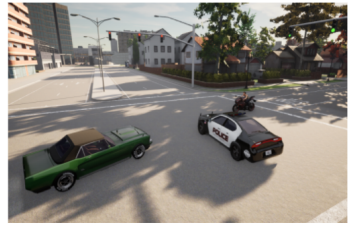

Figure 3. Non-Axiomatic Reasoning System (NARS) logic example. We define background knowledge and the system learns and adapts to new environments and use cases.

## 5. TESTING AND EVALUATION

Several object detection models were compared, including CenterNet [20,21], EfficientDet [22,23], and YOLOv4 [6]. These models were pre-trained on the COCO17 data set [24]. YOLOv4 consistently detected test vehicles with a maximum detection distance of 60.32 m and high confidence of 94% (51.73 m and 45% for CenterNet and 45.38 m and 39% for EfficientDet). YOLOv4 was also the first to detect distant objects. CenterNet detected objects approaching 0.46 seconds later than YOLOv4, and EfficientDet detected objects approaching 0.8 seconds later than YOLOv4.

Then, the YOLOv4 model was further retrained to improve performance on the original COCO17 pretrained model. Increased maximum object detection distance after retraining from 60.32 m to 88.0 m. We also achieved an improved intersection-over-union (IOU) object recognition performance of 0.65 in the adaptive retraining model compared to IOU 0.31 in the COCO data pre-training model. We also cropped the target areas where vehicles are likely to be present. We cropped the image area to 640X640 pixels and used it as the input for the YOLOv4 model, achieving a further distance detection limit of 135 m.

Our object detection system can reliably detect objects about 60-80 meters away from the source. At 60 mph, this gives you about three seconds of warning before a crash. In order to increase the warning time before a crash, the detection range should be increased. To this end, additional sensors such as radar, LIDAR, and depth cameras were applied and performance improvements were evaluated. As a result of comparing the measured truth distance and the radar estimated distance (vertical and horizontal viewing angles of 45 degrees), the errors were 4.7% at 100 m, 6.1% at 200 m, and 8.3% at 300 m. This is a good reference for applications in a simulation environment, but results may vary in real-world situations. Using LIDAR, we obtained distance estimation errors of 5.2% at 100 meters, 7.7% at 200 meters, and 10.3% at 300 meters. We also obtained similar distance estimation error profiles using a depth camera (5.8% for 100 meters, 8.2% for 200 meters, and 11.3% for 300 meters).

In summary, we found radar achieved a minimum distance error distribution compared to LIDAR or depth cameras, so we used radar for object distance calculation and subsequent tracking. Moreover, we can change the angle of the radar field of view, first performing a coarse radar scan of the entire area at a relatively large angle, and then detecting rapidly changing points. It can be used to locate moving objects and narrow the angle of radar scans (fine radar scan) for more precise results.

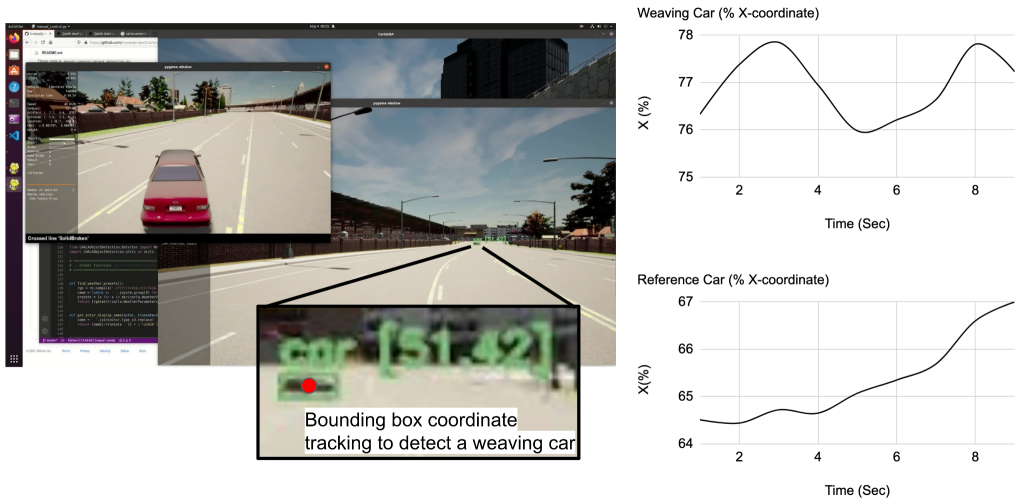

Figure 4. Weave car detection. We detected a car and tracked the x-coordinate movement frame by frame to determine if the car was weaving. We found the weaving from up to 181 meters away.

We detected a weaving car as far as 181 meters away (Figure 4). We also compared the vehicle detection accuracy based on different distances using radar and LIDAR systems. We found that dynamic radar focusing with coarse and fine radar scan achieved better detection accuracy than LIDAR in all distances from 100 to 300 meters. We found that using radar the maximum distance that we detected using radar was 321 meters.

Two scenarios were tested. The first was the risk assessment of the intersection. The system detects approaching vehicles and traffic lights, and NARS sends warning signals based on intersection hazards. In the second scenario, the first responder vehicle is parked on the shoulder of a freeway, and other vehicles approach from behind. The system detected a weaving vehicle approaching from behind and sent an alert 2.7 seconds before the crash (Figure 5).

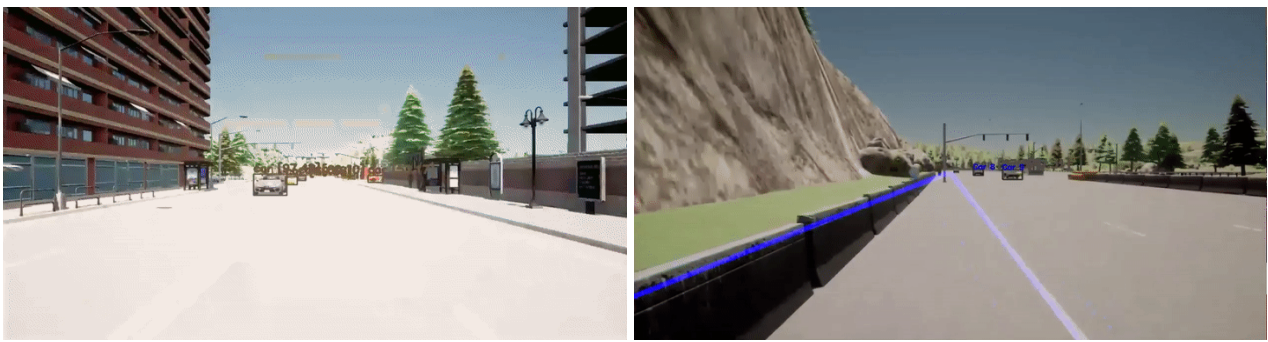

Figure 5. Crossroads risk notification (https://giphy.com/gifs/Ljp4ohbFG336R3PHcJ) and weaving car demonstration (https://giphy.com/gifs/J9Y2sGnpKo7gGYb96s). Other vehicles and traffic lights were recognized correctly. The blue line is the lane detected behind the first responder vehicle. Red text is a NARS notification that has detected an incoming vehicle hazard.

# 6. CONCLUSION

In this study, we showed the improved object recognition performance of our proposed adaptive retraining model compared to the performance of the pre-trained model using COCO data. A radar sensor was used to improve the object detection distance limit in a simulated environment, and deep learning-based object detection and tracking was combined with a neurosymbolic model to demonstrate a weaving car detection function.

The next step is to create multiple sub-scenarios, increasing the variation of each scenario. The transformation points are: Variety of training data can be generated through variations in map location, vehicle presence and model, vehicle location and orientation, weather and lighting, and event timing. We will develop variants in each of these areas, and build and test easy, regular, and difficult versions of each of the scenarios. We will also collect more synthetic video streams and images for training and testing. The data sets to collect include vehicles, people, traffic signs, buildings, lane markings, and lane dividers. Finally, real-world user testing should integrate real-world or virtual-reality driving situations with the head-up display hardware.

# ACKNOWLEDGMENT

The research was carried out at the Jet Propulsion Laboratory, California Institute of Technology, under a contract with the National Aeronautics and Space Administration (80NM0018D0004). The research was funded by the U.S. Department of Transportation, National Highway Traffic Safety Administration - Vehicle Safety Research (DOT-NHTSA) under Task Plan Number 82-106589. We acknowledge the contribution of the system implementation by JPL interns - Daniel Lundstrom, Jacques Joubert, Kevin Yu, Kevin Li, Jessica Chen, Vickie Do, Evelyn Chin, Arya Mevada, Michael Batchev, Peter Isaev, Mina Gabriel, Christian Hahm. We would also like to thank Miami-Dade Police Department George Perera for providing valuable insight regarding the first response operation.